%% file: main.tex
\documentclass[11pt,a4paper]{article}
\usepackage[hyperref]{naaclhlt2019}
\usepackage{times}
\usepackage{latexsym}

\usepackage[utf8]{inputenc} % allow utf-8 input
\usepackage[T1]{fontenc}    % use 8-bit T1 fonts
\usepackage{hyperref}       % hyperlinks
\usepackage{url}            % simple URL typesetting
\usepackage{booktabs}       % professional-quality tables
\usepackage{graphicx}
\usepackage{amsfonts}       % blackboard math symbols
\usepackage{nicefrac}       % compact symbols for 1/2, etc.
\usepackage{microtype}      % microtypography
\usepackage{CJKutf8}
\AtBeginDvi{\input{zhwinfonts}}
\usepackage{amsmath}
\usepackage[linesnumbered,ruled,vlined]{algorithm2e}
\usepackage{multicol}
\usepackage{multirow}

\usepackage{paralist}
\usepackage[american]{babel}
\usepackage{microtype}

\usepackage{url}

\aclfinalcopy % Uncomment this line for the final submission

\title{Multi-Modal Generative Adversarial Network for Short Product Title Generation in Mobile E-Commerce}

\author{
  \textbf{Jian-Guo Zhang,}$^1$ \textbf{Pengcheng Zou,}$^2$ \textbf{Zhao Li,}$^2$  \textbf{Yao Wan,}$^3$  \textbf{Xiuming Pan,}$^2$  \textbf{Yu Gong,}$^2$ \textbf{Philip S. Yu}$^1$ \\
  $^1$Department of Computer Science, University of Illinois at Chicago, Illinois, USA\\
  $^2$Alibaba Group\\
  $^3$College of Computer Science and Technology, Zhejiang University, Hangzhou, China\\
  \texttt{\{jzhan51,psyu\}@uic.edu}, \texttt{wanyao@zju.edu.cn}, \\ \texttt{\{xuanwei.zpc,lizhao.lz,xuming.panxm,gongyu.gy\}@alibaba-inc.com} 
}
\date{}

\begin{document}
\maketitle
\begin{abstract}
  Nowadays, more and more customers browse and purchase products in favor of using mobile E-Commerce Apps such as Taobao and Amazon. Since merchants are usually inclined to describe redundant and over-informative product titles to attract attentions from customers, it is important to concisely display short product titles on limited screen of mobile phones. To address this discrepancy, previous studies mainly consider textual information of long product titles and lacks of human-like view during training and evaluation process. In this paper, we propose a Multi-Modal Generative Adversarial Network (MM-GAN) for short product title generation in E-Commerce, which innovatively incorporates image information and attribute tags from product, as well as textual information from original long titles. MM-GAN poses short title generation as a reinforcement learning process, where the generated titles are evaluated by the discriminator in a human-like view. Extensive experiments on a large-scale E-Commerce dataset demonstrate that our algorithm  outperforms other state-of-the-art methods. Moreover, we deploy our model into a real-world online E-Commerce environment and effectively boost the performance of click through rate and click conversion rate by 1.66\% and 1.87\%, respectively.
  %improves the click through rate by 1.66\% and click conversion rate by 1.87\%, respectively.
  % by a large margin
\end{abstract}

\input{introduction.tex}

\input{related-works.tex}

\input{methodology.tex}
\input{configuration.tex}
\input{experiments.tex}

\input{conclusion.tex}

% \subsubsection*{References}
%\newpage
\bibliographystyle{acl_natbib}
\bibliography{reference}

\end{document}

%% file: introduction.tex
\section{Introduction}
\begin{figure} 
    %\fbox{\rule{0pt}{3.0 in} 
    \centering
    \includegraphics[width=0.88\linewidth]{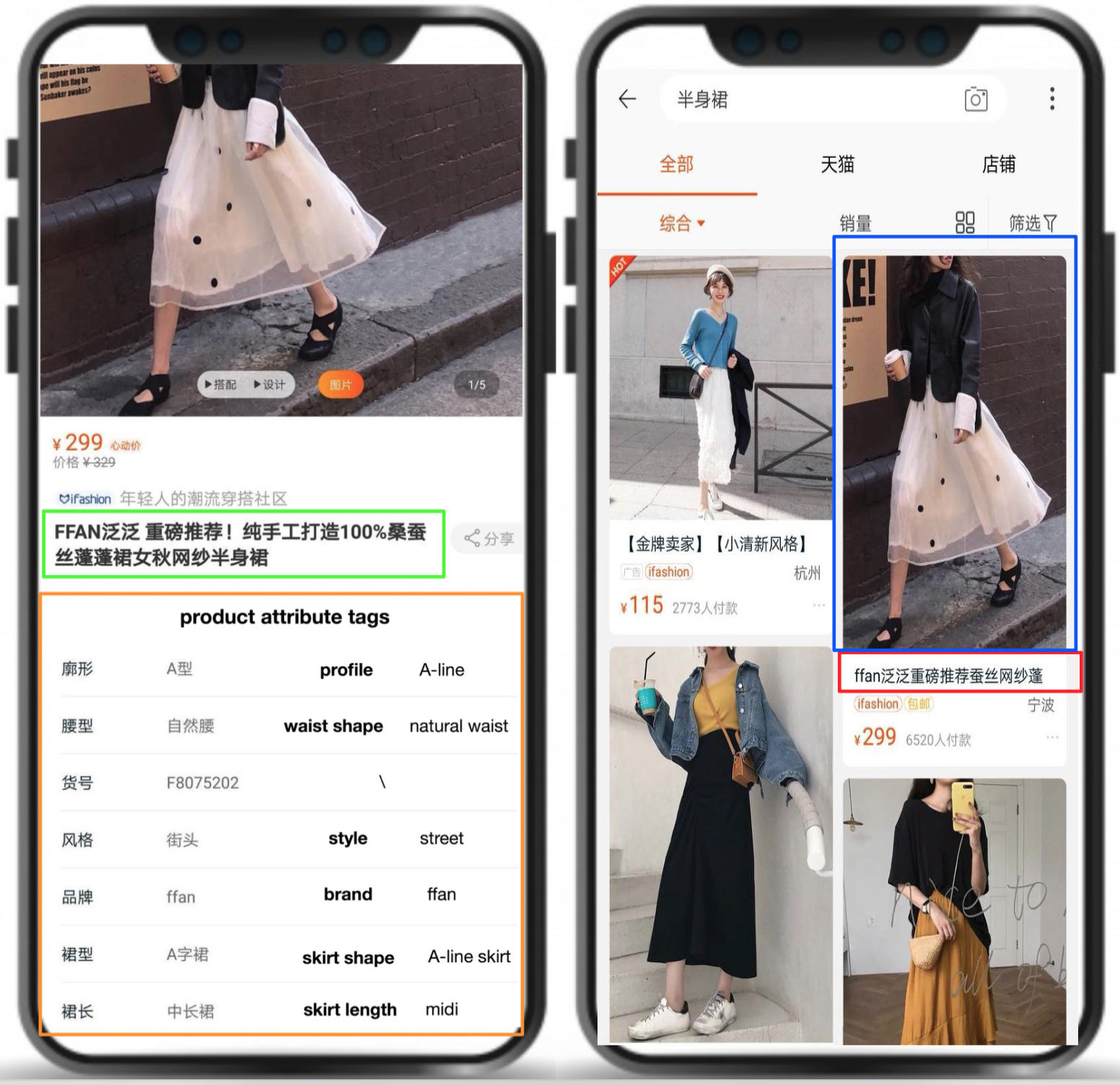}
    %\rule{.9\linewidth}{0pt}}
        \caption{A product with original long titles (green box), cutoff short titles (red box), the main image (blue box), and attribute tags (yellow box).}
        \vspace{-1em}
    \label{fig:example1}
\end{figure}

% E-commerce companies such as TaoBao and Amazon put many efforts to improve the user experience of their apps on cell phones. 
E-commerce companies such as TaoBao and Amazon put many efforts to improve the user experience of their mobile Apps. 
For the sake of improving retrieval results by search engines, merchants usually write lengthy, over-informative, and sometimes incorrect titles, e.g., the original product title in Fig. \ref{fig:example1} contains more than 20 Chinese words, which may be suitable for PCs. However, these titles are cut down and no more than 10 words can be displayed on a mobile phone with limited screen size varying from 4 to 5.8 inches. Hence, to properly display products in mobile screen, it is important to produce succinct short titles to preserve important information of original long titles and accurate descriptions of products.

% On E-commerce platforms, a product is mainly displayed with corresponding image showing visual information, textual title showing key properties, and tags showing attribute information, whereas the textual title is most important to describe a product.
% , whereas a summary or short title is generated from original inputs

This problem is related to text summarization, which can be categorized into two classes: extractive \cite{cao2016attsum,miao2016language,nallapati2017summarunner} and abstractive \cite{chen2016distraction,chopra2016abstractive,see2017get,wan2018improving} methods. The extractive methods select important words from original titles, while the abstractive methods generate titles by extracting words from original titles or generating new words from data corpus. They usually approximate such goals by predicting the next word given previous predicted words using maximum likelihood estimation (MLE) objective. Despite their successes to a large extent, they suffer from the issue of \textit{exposure bias} \cite{ranzato2015sequence}. It may cause the models to behave in undesired ways, e.g., generating repetitive or truncated outputs. In addition, predicting next word based on previously generated words will make the learned model lack of human-like holistic view of the whole generated short product titles.

More recent state-of-the-art methods \cite{gong2018automatic,wang2018multi} treat short product titles generation as a sentence compression task following attention-based extractive mechanism. They extract key characteristics mainly from original long product titles. 
However, in real E-Commerce scenario, product titles are usually redundant and over-informative, and sometimes even inaccurate, e.g., long titles of a cloth may include both
\begin{CJK}{UTF8}{gbsn}
``å»å$|$çé (hip-pop$|$wild)" and ``æèº$|$æ·å¥³ (artsy$|$delicate)"
\end{CJK}
simultaneously. It is a tough task to generate succinct and accurate short titles merely relying on the original titles. Therefore, it is insufficient to regard short title generation as traditional text summarization problem in which original text has already contained complete information.
% Intuitively, we can generate the product description depending on the visual image and the attributes tags of a product. Such that, features of the visual image and attributes of a product are helpful to figure out key words and screen out unrelated words for a short title. 
% In this paper, we employ a multi-modal framework to solve the problem of short title generation, considering not only the original title, but also the image and attribute tags of a product.

% A good product short titles generation model should not produce next key words only based on previously predicted words, it should generate a product short title indistinguishable from a human produced title. Besides, it should also consider other types of information such as the main image and attribute tags of a product. 

% Additionally, a good short title generation model should have a human-like view, and its generated short titles should not only contain a random combination of key words. It should be an eye-catching sentence just like human-written, so that more customers would be willing to click the product. Therefore, abstractive summarization model is promising to generate a better short title. 
% Motivated by adversarial training methods \cite{goodfellow2014generative,yu2017seqgan}, 
In this paper, we propose a novel \textbf{M}ulti-\textbf{M}odal \textbf{G}enerative \textbf{A}dversarial \textbf{N}etwork, named \textbf{MM-GAN}, to better generate short product titles. It contains a generator and a discriminator. The generator generates a short product title based on original long titles, with additional information from the corresponding visual image and attribute tags. On the other hand, the discriminator tries to distinguish whether the generated short titles are human-produced or machine-produced in a human-like view. The task is treated as a reinforcement learning problem, in which the quality of a machine-generated short product title depends on its ability to fool the discriminator into believing it is generated by human, and output of the discriminator is a reward for the generator to improve generated quality. The main contributions of this paper can be summarized as follows:
\begin{compactitem}
\item In this paper, we focus on a fundamental problem existing in the E-Commerce industry, i.e., generating short product titles for mobile E-Commerce Apps. We formulate the problem as a reinforcement learning task;
\item 
% To the best of out knowledge, it's the first attempt that 
We design a multi-modal generative adversarial network to consider multiple modalities 
of inputs for better short product titles generation in E-commerce;
% We propose a novel MM-GAN framework, which incorporates the image and attribute tags aside from original long product titles into the generator. To the best of our knowledge, we give the first attempt to design a multi-modal generative adversarial model to consider multiple modalities of inputs for better short product titles generation in E-commerce;  
\item To verify the effectiveness of our proposed model, we deploy it into a mobile E-commerce App. Extensive experiments on a large-scale real-world dataset with A/B testing show that our proposed model outperforms state-of-the-art methods. 
%\item We propose a novel MM-GAN framework, which incorporates the image and attribute tags aside from original long product titles into the generator. To the best of our knowledge, we give the first attempt to design a multi-modal generative adversarial model to consider multiple modalities of inputs for better short product titles generation in E-commerce;  
% We  propose  a  novel  MM-GAN  framework  for  generat-ing short product titles in E-commerce, it incorporates themain image and attribute tags aside from original productlong titles into the generator
% \item The proposed multi-modal model casts the task as a reinforcement learning problem, where the generated short product titles are evaluated by the discriminator in a human-like view, and return a reward to the generator, pushing it to generate short titles indistinguishable from human-generated titles. 
% The proposed multi-modal model casts the task as a reinforcement learning problem, instead of predicting next word based on previous generated words wiout overall view, where the whole generated product short titles are evaluated by the discriminator, and return a reward to the generator to push it generate short product titles indistinguishable from human-generated titles.
% \item The proposed model outperforms state-of-the-art methods in a large-scale E-commerce dataset. We also deploy the model on an E-commerce mobile app, and see improvement on online sales and conversion rate.
\end{compactitem}

%% file: related-works.tex
\section{Related Work}

Our work is related to text summarization tasks and generative adversarial networks (GANs). 

%\subsection{Text Summarization}
\textbf{Text Summarization.} In terms of text summarization, it mainly includes two categories of approaches: extractive and abstractive methods. Extractive methods produce a text summary by extracting and concatenating several words from original sentence. Whereas abstractive methods generate a text summary based on the original sentence, which usually generate more readable and coherent summaries. Traditional extractive methods such as  graphic models \cite{mihalcea2005language} and optimization-based models \cite{woodsend2010automatic} usually rely on human-engineered features. Recent RNN-based methods \cite{chopra2016abstractive,gong2018automatic,nallapati2017summarunner,wang2018multi} have become popular in text summarization tasks. 
Among them, \cite{gong2018automatic,wang2018multi} design attention-based neural networks for short product titles generation in E-commerce. \cite{gong2018automatic} considers rich semantic features of long product titles. \cite{wang2018multi} designs a multi-task model and uses user searching log data as additional task to facilitate key words extraction from original long titles. However, they mainly consider information from textual long product titles, which sometimes are not enough to select important words and filter out over-informative and irrelevant words from long product titles. In addition, these methods mainly apply MLE objective and predict next word based on previous words. As a consequence, these models usually suffer from \textit{exposure bias} and lack of human-like view.

\textbf{Generative Adversarial Networks (GANs)}.
GAN is firstly proposed in \cite{goodfellow2014generative}, which is designed for generating real-valued, continuous data, and has gained great success in computer vision tasks \cite{dai2017towards,isola2017image,ledig2017photo}.  However, applying GANs to discrete scenarios like natural language has encountered many difficulties, since the discrete elements break the differentiability of GANs and prohibit gradients backpropagating from the discriminator to generator. 
% Due to this reason, GAN has not achieved comparable success in NLP. Recently, researchers usually reformulate the problem to continuous domain or consider reinforcement learning problems. 
To mitigate these above mentioned issues, SeqGan \cite{yu2017seqgan} models sequence generation procedure as a sequential decision making process. It applies a policy gradient method to train the generator and discriminator, and shows improvements on multiple generation task such as poem generation and music generation. MaskGan \cite{fedus2018maskgan} designs a actor-critic reinforcement learning based GAN to improve qualities of text generation through filling in missing texts conditioned on the surrounding context. There are also some other RL based GANs for text generation such as LeakGan \cite{guo2017long}, RankGan \cite{lin2017adversarial}, SentiGan \cite{wang2018sentigan}, etc. All above methods are designed for unsupervised text generation tasks. 
\cite{li2017adversarial} designs an adversarial learning method for neural dialogue generation. They train a seq2seq model to produce responses and use a discriminator to distinguish whether the responses are human-generated and machine-generated, and showing promising results.
% They also propose some methods on how to stabilize training for GANs such as pretraining method. In this paper, we adopt their pretraining method before normal training process.  
It should be noticed that our work differs from other similar tasks such as image captioning \cite{dai2017towards} and visual question answering \cite{antol2015vqa}. The image captioning can be seen as generating caption from a single modality of input, while the visual question answering mainly focuses on aligning the input image and question to generate a correct answer. In our task, we put more attention on learning more information from the multi-modal input sources (i.e., long product titles, product image and attribute tags) to generate a short product title.

%% file: methodology.tex
\section{Multi-Modal Generative Adversarial Network}
% \begin{figure*} 
%     %\fbox{\rule{0pt}{3.0 in}
%     % \resizebox{1.0\textwidth}{!}{%
%     \centering
%     \includegraphics[width=0.82\textwidth]{images/framework.eps}
%     %\rule{.9\linewidth}{0pt}}
%     \caption{Overall Framework of MM-GAN.}
%     %}å
%     \label{fig:framework1}
% \end{figure*}

% $O=\left \{o_1, o_2, ..., o_K\right \}$
% $R= \left \{r_1, r_2, ...,r_N\right \}$$
% $A=\left [a_1, a_2, ..., a_M \right]$
% $L=\left [l_1, l_2, ..., l_K \right]$
% $U=\left \{u_1, u_2, ..., u_M\right \}$

In this section, we describe in details the proposed MM-GAN. The problem can be formulated as follows: given an original long product title $L=\left \{l_1, l_2, ..., l_K \right\}$
consisted of $K$ Chinese or English words, a single word can be represented in a form like ``skirt" in English or \begin{CJK}{UTF8}{gbsn}``半身裙"\end{CJK} in Chinese. With an additional image $I$ and attribute tags $A=\left \{a_1, a_2, ..., a_M \right\}$, the model targets at generating a human-like short product title $S= \left \{s_1, s_2, ...,s_N\right \}$, where $M$ and $N$ are the number of words in $A$ and $S$, respectively.  
% The process of short titles generation can be treated as a sequence of actions that are taken according to a policy defined by the multi-modal generator. The discriminator distinguish whether the generated short product titles are human-generated or machine-generated, and return a reward to the multi-modal generator.
% The process of short titles generation can be treated as a sequence of actions that are taken according to a policy defined by the multi-modal generator. The discriminator distinguish whether the generated short product titles are human-generated or machine-generated, and return a reward to the multi-modal generator.

\subsection{Multi-Modal Generator}
\begin{figure} 
    %\fbox{\rule{0pt}{3.0 in}
    % \resizebox{1.0\textwidth}{!}{%
    \centering
    \includegraphics[width=0.48\textwidth]{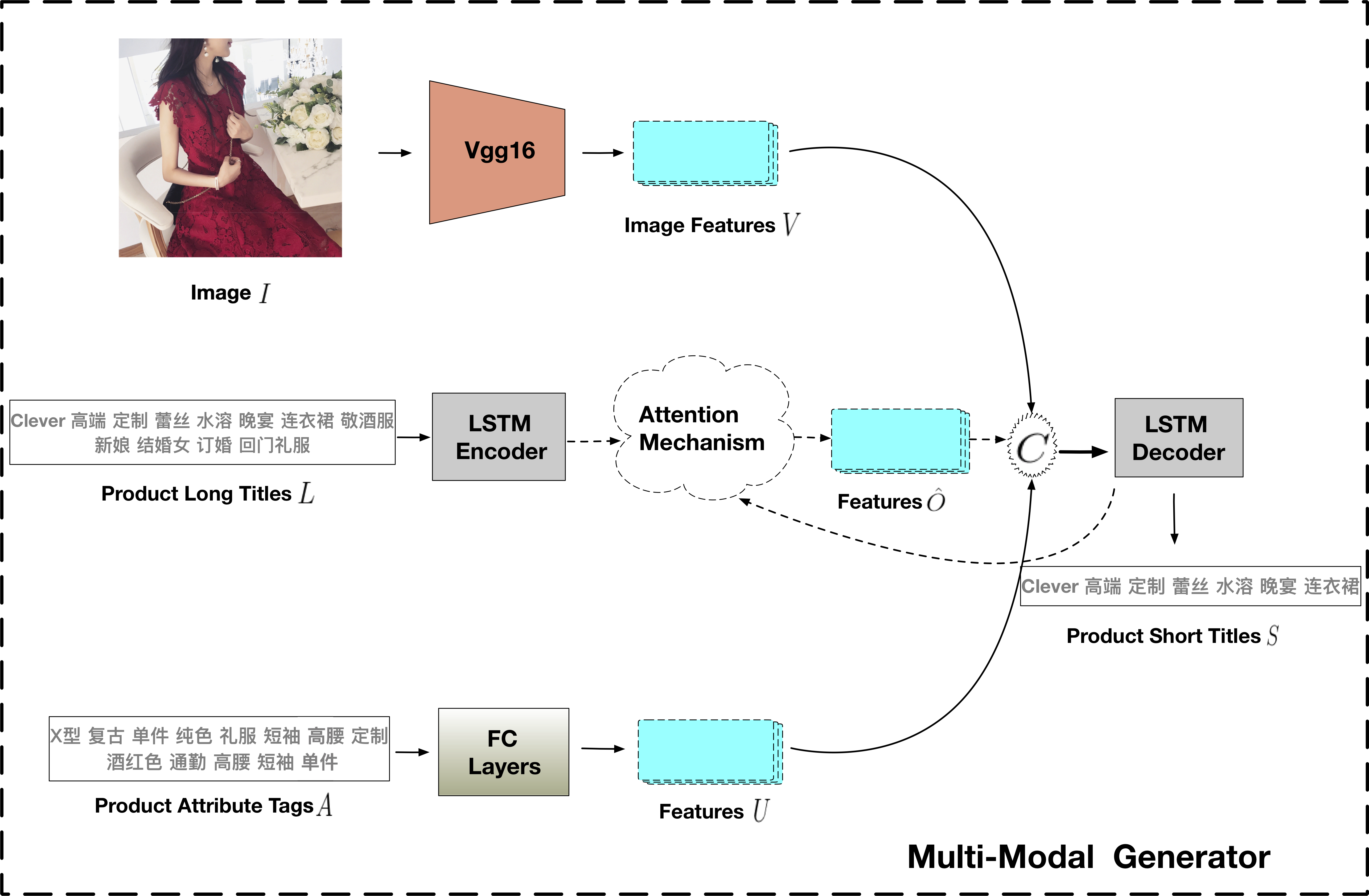}
    %\rule{.9\linewidth}{0pt}}
    \caption{Overall Framework of MM-GAN.}
    %}å
    \label{fig:framework1}
\end{figure}

The multi-modal generative model defines a policy of generating short product titles $S$ given original long titles $L$, with additional information from product image $I$ and attribute tags $A$. Fig. \ref{fig:framework1} illustrates the architecture of our proposed multi-modal generator which follows the seq2seq \cite{sutskever2014sequence} framework.
%\subsubsection{Multi-Modal Encoder}
% \begin{wrapfigure}{r}{0.5\textwidth}
%     %\fbox{\rule{0pt}{3.0 in}
%     % \resizebox{1.0\textwidth}{!}{%
%     \centering
%     \includegraphics[width=0.5\textwidth]{images/nips-w.eps}
%     %\rule{.9\linewidth}{0pt}}
%     \caption{Multi-Modal Generator of MM-GAN.}
%     %}å
%     \label{fig:framework1}
% \end{wrapfigure}

\textbf{Multi-Modal Encoder. }As we mentioned before, our model tries to incorporate multiple modalities of a product (i.e., product image, attribute tags and long title). To learn the multi-modal embedding of a product, we first adopt a pre-trained VGG16 \cite{simonyan2014very} as the CNN architecture to extract features $V=[v_1, v_2, ..., v_Z]$ of an image $I$ from the condensed fully connected layers, where $Z$ is the number of latent features. In order to get more descriptive features, we fine-tune the last 3 layers of VGG16 based on a supervised classification task given classes of products images. 
% Note here we don't use spatial features from convolutional neural network as Image Caption and Visual Questioning Tasks \cite{dai2017towards,shetty2017speaking,wu2017you}, in which image is most important part.  
% To incorporate different sources of textual inputs, we employ a seq2seq model \cite{sutskever2014sequence} as the basic architecture. 
Second, we encode the attribute tags $A$ into a fixed-length feature vector $U=\left [u_1, u_2, \ldots, u_{M'}\right ]$, and $U=f_1(A)$, where $f_1$ denotes fully connected layers, $M'$ is the output size of $f_1$.
Third, we apply a recurrent neural network to encode the original long titles $L$ as $O=\left [o_1, o_2, \ldots, o_K \right]$, where $o_t=f_2(o_{t-1}, l_t)$. Here $f_2$ represents a non-linear function, and in this paper the LSTM unit \cite{hochreiter1997long} is adopted.
% which is the hidden state of a separate non-linear function $f_2$ at step $t$ given input word $l_t$, and $f_2$ are LSTM \cite{hochreiter1997long} cells with different layers.

\textbf{Decoder.}
The hidden state $h_t$ for the $t$-th target word $s_t$ in short product titles $S$ can be calculated as $h_t=f_2(h_{t-1},s_{t-1}, \hat{o}_t)$. Here we adopt an attention mechanism \cite{bahdanau2014neural} to capture important words from original long titles $L$. The context vector $\hat{o}_t$ is a weighted sum of hidden states $O$, which is represented as:
% {\small
\begin{equation}
    \hat{o}_t=\sum_{k=1}^{K}\alpha_{t,k}o_k,
\end{equation}
%}
where $\alpha_{t,k}$ is the contribution of an input word $l_k$ to the $t$-th target word using an alignment model $g$ \cite{bahdanau2014neural}:
% {\small
\begin{equation}
    \alpha_{t,k}=\frac{\exp(g(h_{t-1}, o_k))}{\sum_{k'=1}^{K}\exp(g(h_{t-1}, o_{k'}))}.
\end{equation}
% }

After obtaining all features $U$, $V$, $\hat{O}$ from $A$, $I$ and $L$, respectively, we then concatenate them into the final feature vector:

% {\small
\begin{equation}
  C=\tanh(W[\hat{O};V;U]),  
\end{equation}
% }
where $W$ are learnable weights and $[;]$ denotes the concatenation operator. 

Finally, $C$ is fed into the LSTM based decoder to predict the probability of generating each target word for short product titles $S$. As the sequence generation problem can be viewed as a sequence decision making process \cite{bachman2015data}, we denote the whole generation process as a policy $\pi(S|C)$.

\subsection{Discriminator}
The discriminator model $D$ is a binary classifier which takes an input of a generated short product titles $S$ and distinguishes whether it is human-generated or machine-generated. The short product titles are encoded into a vector representation through a two-layer LSTM model, and then fed into a sigmoid function, which returns the probability of the input short product titles being generated by human:
% {\small
\begin{equation}
   R_{\phi }(S)=sigmoid(W_d\left [ LSTM(S)\right ]+b_d),
\end{equation}
% }
where $\phi$ are learnable parameters for D, $W_d$ is a weight matrix and $b_d$ is a bias vector.  

% \subsection{Objectives and Training for MM-GAN}
\subsection{End-to-End Training}
The multi-modal generator $G$ tries to generate a sequence of tokens $S$ under a policy $\pi$ and fool the discriminator $D$ via maximizing the reward signal received from $D$. The objective of $G$ can be formulated as follows:

% {\small
\begin{equation}
    J(\theta)=\mathbb{E}_{S\sim \pi_{\theta}\left ( S|C \right )}\left [ R_{\phi }(S) \right ],
\end{equation}
% }
where $\theta$ are learnable parameters for $G$. 
%Given the input long product titles $L$ with additional image $I$ and attribute tags $A$, $G$ generates a short product titles $S$ by sampling from the policy. 

Conventionally, GANs are designed for generating continuous data and thus $G$ is differentiable with continuous parameters guided by the objective function from $D$ \cite{yu2017seqgan}. Unfortunately, it has difficulty in updating parameters through back-propagation when dealing with discrete data in text generation. To solve the problem, we adopt the REINFORCE algorithm \cite{williams1992simple}. Specifically, once the generator reaches the end of a sequence (i.e., $S=S_{1:T}$), it receives a reward $R_{\phi}\left ( S \right )$ from $D$ based on the probability of being real. 

%In reinforcement learning, the current reward is not only determined by previous intermediate states, but also compromised by future outcome states. However,

In text generation, $D$ will provide a reward to $G$ only when the whole sequence has been generated, and no intermediate reward is obtained before the final token of $S$ is generated. This may cause the discriminator to assign a low reward to all tokens in the sequence though some tokens are proper results. 
To mitigate the issue, we evaluate the reward function by aggregating the $N'$-time Monte-Carlo simulations \cite{yu2017seqgan} at each decoding step:
% {\small
%\begin{equation}
\begin{align}
&R_{\phi}^{'}(S_{1:t-1}, a'=s_t) \approx \\ & \left\{\begin{matrix}\frac{1}{N'}\sum_{n=1}^{N'}R(S_{1:t}, S_{t+1:N}^{(n)}),\   t< N
\\ R(S_{1:t-1}, s_t), \ \ \ \ \ \ \ \ \ \ \ \ \ \ \ \ \ \ \ \ \ \ \ t=N,
\end{matrix}\right. \nonumber
\end{align}
%\end{equation}
% }
where $ \{ S_{t+1:N}^{(1)}, \ldots, S_{t+1:N}^{(N')}  \}$ is the set of generated short titles, which are sampled from the $t+1$-th decoding step based current state and action.
% \begin{align}
%     R_{\phi}(S_{1:t-1}, a'=s_t)=\frac{1}{N'}\sum_{n=1}^{N'}R(S_{1:N}^n)
% \end{align}
% we utilize Monte Carlo (MC) search with $N'$-time roll-outs \cite{yu2017seqgan} to assign rewards to intermediate tokens.  
% $\left \{ S_{1:N}^1, \ldots, S_{1:N}^{N'} \right \}=MC^{\pi}\left ( S_{1:N}; N'\right )$
% where $S_{1:t}^n=\left ( s_1, \ldots, s_t \right )$ and $S_{t+1:T}^n$ are sampled based on roll-out policy $\pi$ and the current state. The intermediate reward now is $R_{\phi}(S_{1:t-1}, a'=s_t)=\frac{1}{N'}\sum_{n=1}^{N'}R(S_{1:N}^n)$,
% here $a'$ is an action at current state $t$. 
Now we can compute the gradient of the objective function for the generator $G$:
% {\small
\begin{align} \label{eq-generator} 
       &\nabla_{\theta} J(\theta) \approx \\ 
    %   & \approx \mathbb{E}_{S\sim \pi_{\theta}\left ( S|C \right )}\left [\pi_{\theta \left ( S|C \right )}\nabla_{\theta} log\left ( \pi_{\theta \left ( S|C \right )} \right ) R_{\phi }(S) \right ], \\ \nonumber 
      & \mathbb{E}_{S\sim \pi_{\theta}\left ( S|C \right )}\left [ \sum_{t=1}^{N} \nabla_{\theta}\log(\pi_{\theta\left ( s_t|C \right )})R_{\phi}^{'}(S_{1:t-1}, s_t)\right ], \nonumber
\end{align}
% }
where $\nabla_{\theta}$ is the partial differentiable operator for $\theta$ in $G$, and the reward $R_{\phi}^{'}$ is fixed during updating of generator. 

The objective function for the discriminator $D$ can be formulated as:
 {\small
\begin{equation}\label{object-function} 
    \mathbb{E}_{S\sim \mathcal{P}_{\theta}\left ( S|C \right )}\left [ \log R_{\phi}^{'}(S|C) \right ]-\mathbb{E}_{S\sim \pi_{\theta}\left ( S|C \right )}\left [ \log R_{\phi}^{'}(S|C) \right ],
\end{equation}
}
where $S \sim \mathcal{P}_{\theta}$ and $S \sim \pi_{\theta}$ denote that $S$ is from human-written sentences and synthesized sentences, respectively.
%======================================================
\begin{algorithm}
	\SetAlgoVlined
	\small
	%\footnotesize
	%\scriptsize
	\KwIn{ Long product titles $L$, short titles $S$, images $I$, attribute tags $A$. Multi-modal generator $G$, discriminator $D$.}
	Fine-tune last 3 layers of pretrained VGG16 network to get image features $V$ based on a classification task \\
	Pretrain $G$ on human-generated data using MLE\\
	Pretrain $D$ on human-generated data and machine-generated data\\
	\For {number of training iterations}{
		%  \For{$i\gets0$ \KwTo $B-1$}{
		\For{$i \gets1$ \KwTo D-steps}{
			Sample $S$ from human-generated data \\
			Sample $\hat{S}\sim \pi_{\theta}(\cdot|C)$ \\
			Update $D$ on both $S$ and $\hat{S}$ 
		}
		\For{$i \gets1$ \KwTo G-steps}{
			Sample $(L,V,A,S)$ from human-generated data \\
			Sample $(L,V,A,\hat{S})\sim \pi_{\theta}(\cdot|C)$ based on MC search\\
			Compute reward $R_{\phi}^{'}$ for $(L,V,A,\hat{S})$ using $D$\\
			Update $G$ using $R_{\phi}^{'}$ based on Eq. \eqref{eq-generator}\\
			Teacher-Forcing: Update $G$ on $(L,V,A,S)$ using MLE
		}
	}
	\small\caption{Multi-Modal Generative Adversarial Network}
	\label{alg:1}
\end{algorithm}

On training stage, we first pre-train $G$ and $D$ for several steps. Due to the large size of searching space of possible sequences, it is also very important to feed human-generated short product titles to the generator for model updates. Specifically, we follow the Teacher Forcing mechanism \cite{lamb2016professor}. In each training step, we first update the generator using machine-generated data with rewards gained from the discriminator, then sample some data from human-generated short product titles and assign a reward of $1$ to them, where the generator uses this reward to update parameters. Alg. \ref{alg:1} summarizes the training procedure, where D-steps and G-steps are both set to 1 in this paper.

%====================================================

%% file: configuration.tex
\section{Experiments}
% In order to empirically evaluate the effectiveness of the proposed method in generating short product titles on E-commerce, we conduct experiments on a large scale data set and compare it with state-of-the-art methods. We further implement the framework in a real world online environment to test its practical performance. 

\subsection{Experimental Setup}

\textbf{Dataset. }
The dataset used in our experiment is crawled from a module named \begin{CJK}{UTF8}{gbsn}有好货 (Youhaohuo)\end{CJK} of the well-known \begin{CJK}{UTF8}{gbsn}淘宝 (TAOBAO)\end{CJK} platform in China. Every product in the dataset includes a long product title and a short product title written by professional writers, along with product several high quality visual images and attributes tags, here for each product we use its main image. This Youhaohuo module includes more than $100$ categories of products, we crawled top $7$ categories of them in the module, and exclude the products with original long titles shorter than 10 Chinese characters. 
% since product titles shorter than 10 Chinese characters can be completely displayed in most mobile screens (Wang et al. 2018).
We further tokenize the original long tittles and short titles into Chinese or English words, e.g. “skirt” is a word in English and \begin{CJK}{UTF8}{gbsn}半身裙\end{CJK}  is a word in Chinese. Table \ref{tab:table11} shows some details of the dataset. We randomly select $1.6$M samples for training, $0.2$M samples for validation, and test our proposed model on $5000$ samples.

\begin{table} 
\centering
\resizebox{0.8\linewidth}{!}{%
\begin{tabular}{|l|l|} 
\hline
Data  set size                 & 2,403,691 \\ \hline
Avg. length of long Titles & 13.7 \\ \hline
Avg. length of Short Titles    & 4.5 \\ \hline
Avg. length of Attributes Tags & 18.3 \\ \hline
Avg. number of Image      & 1 \\ \hline
\end{tabular}
}
\caption{Statistics of the crawled dataset. Here all the lengths are counted by Chinese or English words.} \label{tab:table11} 
\end{table}

\textbf{Baselines.} We compare our proposed model with the following four baselines: (a) Pointer Network (\textbf{Ptr-Net}) \cite{see2017get} which is a seq2seq based framework with pointer-generator network copying words from the source text via \textit{pointing}.
% : It is an attention based seq2seq framework, which leverage a pointer-generator architecture with coverage to generate short summary based on original long texts. 
(b) Feature-Enriched-Net (\textbf{FE-Net}) \cite{gong2018automatic} which is a deep and wide model based on attentive RNN to generate the textual long product titles. 
(c) Agreement-based MTL (\textbf{Agree-MTL}) \cite{wang2018multi} which is a multi-task learning approach to improve product title compression with user searching log data.
% , where the agree-based loss is interpolated into the original loss function during training 
(d) Generative Adversarial Network (\textbf{GAN}) \cite{li2017adversarial} which is a generative adversarial method for text generation with only one modality of input. %For these baselines, we follow their initial setting, tune parameters on validation set and report their best results on test set. 

\textbf{Implementation Details.}
%\subsection{Implementation details} %standard maximum likelihood training
% Before training, we first perform pretraining. 
% To train a stable model, 
We first pre-train the multi-modal generative model given human-generated data via maximum likelihood estimation (MLE), and we transfer the pretrained model weights for the multi-modal encoder and decoder modules. Then we pre-train the discriminator using human-generated data and machine-generated data. To get training samples for the discriminator, we sample half of data from multi-modal generator and another half from human-generated data. After that, we perform normal training process based on pre-trained MM-GAN.

Specifically, we create a vocabulary of 35K words for long product titles and short titles, and another vocabulary of 35k for attribute tags in training data with size of $1.6$M. Words appear less than $8$ times in the training set are replaced as $<$UNK$>$. We implement a two-layer LSTM with $100$ hidden states to encoder attribute tags, and all other LSTMs in our model are two layers with $512$ hidden states. The last $3$ layers of the pre-trained VGG16 network are fine tuned based on the products visual images with $7$ classes. The Adam optimizer \cite{kingma2014adam} is initialized with a learning rate $10^{-3}$. The multi-modal generator and  discriminator are pre-trained for $10000$ steps, the normal training steps are set to $13000$, the batch size is set to $512$ for the discriminator and $256$ for the generator, the MC search time is set to $7$. 
% All models are trained and tested on a single NVIDIA Tesla P100 GPU with 16GB memory.  

%% file: experiments.tex
\begin{table*}
\centering
\resizebox{0.8\textwidth}{!}{%
\begin{tabular}{l|cccccc}
\hline
\textbf{Models}     & \textbf{Ptr-Net} & \textbf{FE-Net} & \textbf{Agree-MTL} & \textbf{GAN} & \textbf{MM-GAN} \\ \hline
ROUGE-1  & 59.21   & 61.07  & 66.19     & 60.67       &\textbf{69.53}    \\
ROUGE-2   & 42.01   & 44.16  & 49.39     & 46.46       &\textbf{52.38}  \\
ROUGE-L & 57.12   & 58.00  & 64.04     & 60.27       &\textbf{65.80} \\
\hline
\end{tabular}
}
\caption{ROUGE performance of different models on the test set.}
\label{tab-table2}
\end{table*}

% Please add the following required packages to your document preamble:
% \usepackage{multirow}
\begin{table*}[]
\resizebox{\textwidth}{!}{%
\begin{tabular}{|cc|l|l|}
\hline
\multicolumn{2}{|c|}{Data}                                          & \multicolumn{1}{c|}{Methods} & \multicolumn{1}{c|}{Results} \\ \hline
Product Long Titles                         & \multicolumn{1}{l|}{\begin{tabular}[c]{@{}l@{}} \begin{CJK}{UTF8}{gbsn}Artka 阿卡 夏新 花边 镂空 荷叶边 抽绳 民族 狂野 复古 衬衫 S110061Q\end{CJK}\\ (Artka Artka summer lace hollow-out flounce drawstring \\ nation wild retro shirt S110061Q) \end{tabular} } & FE-Net                       &      \begin{tabular}[c]{@{}l@{}} \begin{CJK}{UTF8}{gbsn}阿卡 花边 镂空 荷叶边  衬衫\end{CJK} \\ (Artka lace hollow-out flounce  shirt) \end{tabular}                            \\ \hline
\multicolumn{1}{|c|}{\multirow{3}{*}{Image \begin{minipage}{0.3\textwidth}
      \includegraphics[width=0.5\linewidth]{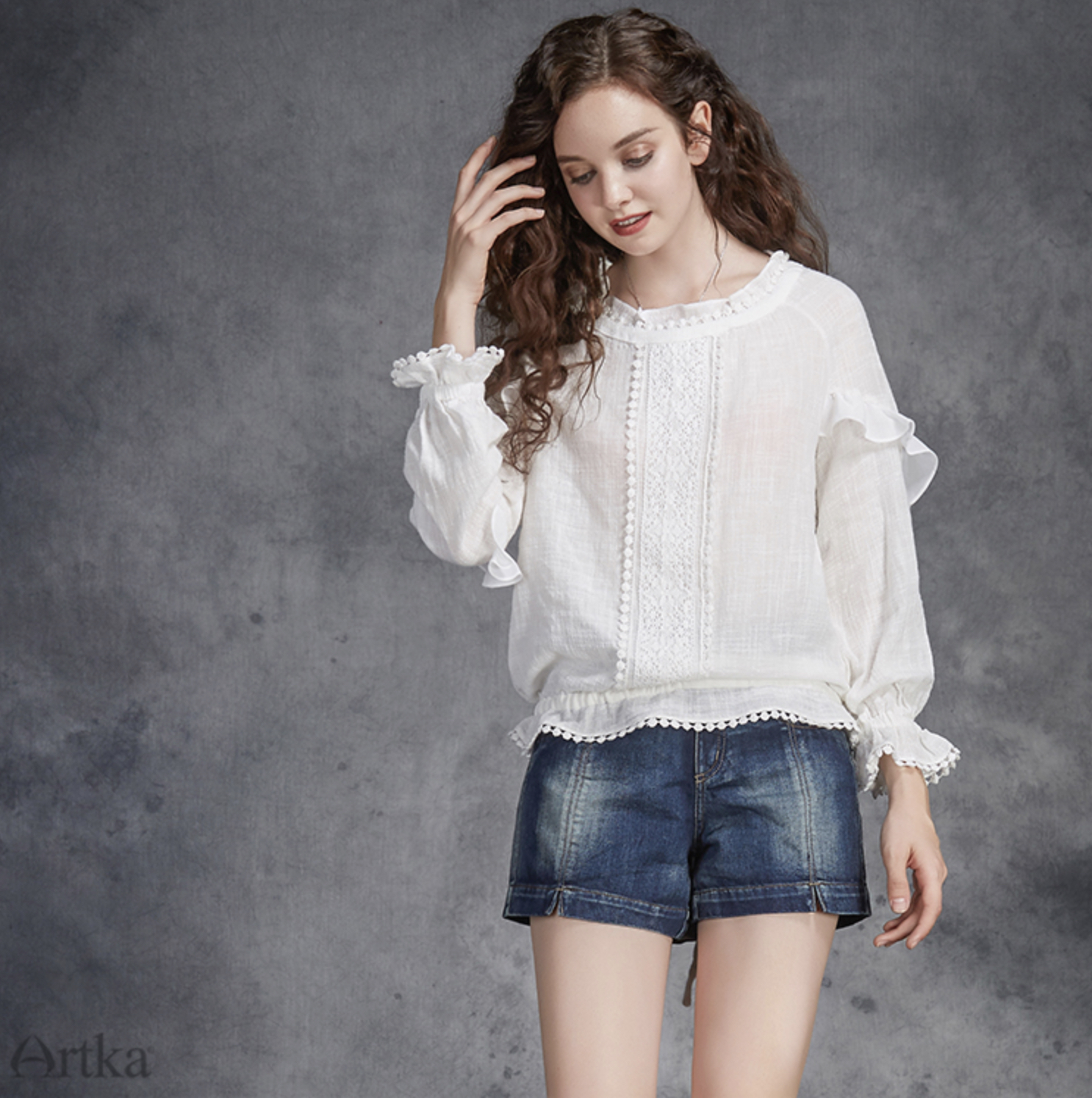}
    \end{minipage}}} & \multirow{3}{*}{Attributes Tags  \begin{tabular}[c]{@{}l@{}} \begin{CJK}{UTF8}{gbsn}修身 常规款 圆领  Artka 米白 长袖 \end{CJK} \\ \begin{CJK}{UTF8}{gbsn}套头 复古  通勤 纯色 夏季 喇叭袖 棉\end{CJK}  \\ (slim common round-neck Artka off-white long-sleeve \\ pullover retro commuting plain summer flare-sleeve cotton)\end{tabular} }     & Agree-MTL                    &             \begin{tabular}[c]{@{}l@{}} \begin{CJK}{UTF8}{gbsn}Artka 阿卡 夏新 花边 镂空 荷叶边 衬衫\end{CJK} \\ (Artka Artka summer lace hollow-out flounce  shirt) \end{tabular}                    \\ \cline{3-4} 
\multicolumn{1}{|c|}{}                      &                       & GAN                  &                          \multicolumn{1}{l|}{\begin{tabular}[c]{@{}l@{}} \begin{CJK}{UTF8}{gbsn}Artka 荷叶边 抽绳 衬衫\end{CJK} \\ (Artka  lace  flounce drawstring shirt) \end{tabular} }                \\ \cline{3-4} 
\multicolumn{1}{|c|}{}                      &                       & MM-GAN                       &               \multicolumn{1}{l|}{\begin{tabular}[c]{@{}l@{}} \begin{CJK}{UTF8}{gbsn}Artka 花边 荷叶边 镂空 复古 衬衫\end{CJK} \\ (Artka  lace flounce hollow-out retro shirt) \end{tabular}}                 \\ \hline
\end{tabular}
}
\caption{The comparison of generated short titles among different methods.} \label{tab-case}
\end{table*}

\subsection{Automatic Evaluation}
To evaluate the quality of generated product short titles, we follow \cite{wang2018multi,gong2018automatic} and use standard recall-oriented ROUGE metric \cite{lin2004rouge}, which measures the generated quality by counting the overlap of N-grams between the machine-generated and human- generated titles. Here we consider ROUGE-1 (1-gram), ROUGE-2 (bi-grams), ROUGE-L (longest common subsequence).
% ROUGE measures the generated quality by counting the overlap of N-grams between the machine-generated and human-generated titles. Here we consider ROUGE-1 ($1$-gram), ROUGE-2 (bi-grams), ROUGE-L (longest common subsequence). 
Experimental results on the test set are shown in Table \ref{tab-table2}. From this table, we note that our proposed MM-GAN achieves best performance on three metrics. 
% Comparing with the best model Agree-MTL, MM-GAN improves  ROUGE-1, ROUGE-2, ROUGE-L
% by $3.34\%$, $2.99\%$, $1.76\%$, respectively. 
Furthermore, when comparing MM-GAN with GAN,
we can see that our proposed MM-GAN achieves an improvement of  $8.86\%$, $5.92\%$, $5.53\%$, in terms of ROUGE-1, ROUGE-2, ROUGE-3, respectively. This verifies that additional information such as image and attribute tags from product can absolutely facilitate our model to generate better short titles. 
In addition, compared with the best model Agree-MTL, MM-GAN improves ROUGE-1, ROUGE-2, ROUGE-L
by $3.34\%$, $2.99\%$, $1.76\%$, respectively. We attribute the outperformance of MM-GAN to two facts: 
(a) it incorporates multiple sources, containing more information than other single-source based models.
(b) it applies a discriminator to distinguish whether a product short titles are human-generated or machine-generated, which makes the model evaluate the generated sequence in a human-like view, and naturally avoid exposure bias in other methods. 

% %%%%%%%%%%%%%%%%%%%%%%%%%%%%%

\subsection{Online A/B Testing}
% In order to further verify the effectiveness of MM-GAN, we test our method in the real world online environment of an E-commerce app.
% We have verified effectiveness of the proposed method in previous experiments. 
In order to further verify the effectiveness of MM-GAN, we test our method in the real-world online environment of the TaoBao App.

We perform A/B testing in seven categories of products in the E-commerce App, i.e.,  \begin{CJK}{UTF8}{gbsn}``连衣裙$|$ (one-piece)", ``男士T恤$|$ (Man T-shirt), ``衬衫$|$ (shirt)'', ``休闲裤$|$ (Casual pants)", ``女士T恤 $|$ (Woman T-shirt)'', ``半身裙$|$ (skirt)", ``毛针织衫$|$ (Sweaters)"\end{CJK}. During online A/B testing,  users ($3\times 10^6$ users per day) are split equally into two groups and are directed into a baseline bucket and an experimental bucket. For users in the baseline bucket, product short titles are generated by the default online system, following an ILP based method \cite{clarke2008global}. While for users in the experimental bucket, product short titles are generated by MM-GAN. All conditions in the two buckets are identical except for short titles generation methods. We apply Click Through Rate (CTR) and Click Conversion Rate (CVR) to measure the performance. $CTR=\frac{\#click\_of\_product}{\#pv\_of\_product}$, and $ CVR=\frac{\#trade\_of\_product}{\#click\_of\_product}$, where $\#click\_of\_product$ indicates clicking times of a product, $\#pv\_of\_product$ is the number of page views of the product and $\#trade\_of\_product$ is the number of purchases of the product.

We deploy A/B testing for $7$ days and calculate overall CTR for all products in the baseline bucket and experimental bucket. MM-GAN improves CTR by $1.66\%$ and CVR by $1.87\%$ in the experimental bucket over the default online system in the baseline bucket. It verifies the effectiveness of our proposed method. Moreover, through online A/B testing, we find that with more readable, informative product short titles, users are more likely to click, view and purchase corresponding products, which indicates good short product titles can help improving the product sales on E-commerce Apps.    

\subsection{Qualitative Analysis}

\begin{figure*} 
    %\fbox{\rule{0pt}{3.0 in}
    % \resizebox{0.8\textwidth}{!}{%
    \centering
    \includegraphics[width=0.93\textwidth]{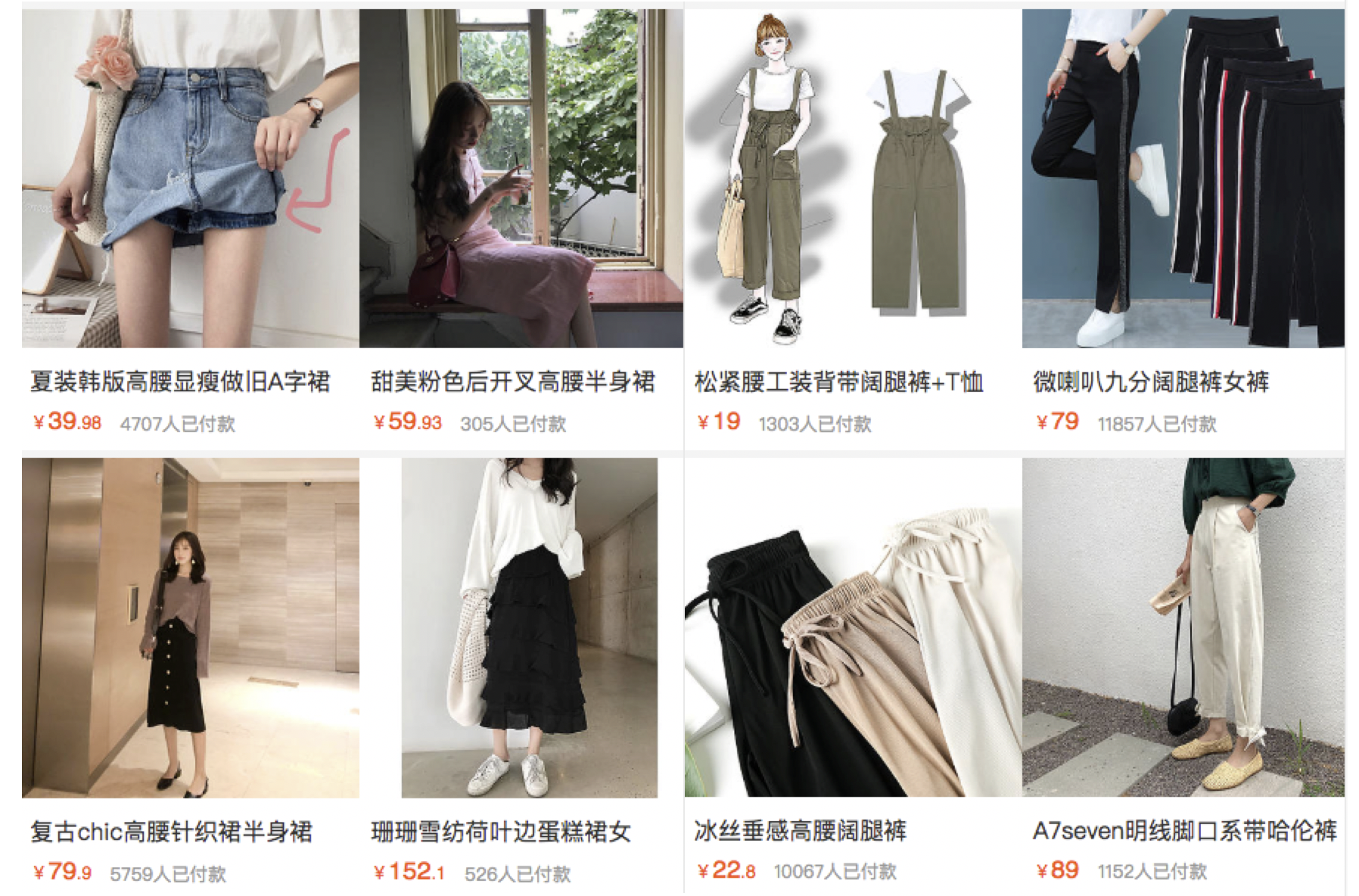}
    %\rule{.9\linewidth}{0pt}}
    \caption{Some samples generated by MM-GAN.}
   % }
    \label{fig:example-demo}
\end{figure*}

Table \ref{tab-case} shows a sample of product short titles generated by MM-GAN and baselines.
% \footnote{More sampes can be found in the supplementary material.}
 
From this table, we can note that (a) product short titles generated by our model are more fluent, informative than baselines, and core product words (e.g., \begin{CJK}{UTF8}{gbsn}``Artka$|$ (阿卡)", ``复古$|$ (retro)", ``衬衫$|$ (shirt)"\end{CJK}) can be recognized. (b) There are over-informative words (e.g., \begin{CJK}{UTF8}{gbsn}``阿卡$|$ (Artka)", ``S110061Q"\end{CJK}) and irrelevant words (e.g., \begin{CJK}{UTF8}{gbsn}``狂野$|$ (wild)"\end{CJK}) in product long titles. Over-informative words may disturb model's generation process, irrelevant words may give incorrect information to the model. These situations could happen in real E-commerce environment. FE-Net misses the English brand name ``Artka" and gives its Chinese name \begin{CJK}{UTF8}{gbsn}`阿卡"\end{CJK} instead. Agree-MTL using user searching log data performs better than GAN. However, Agree-MTL still generates the over-informative word \begin{CJK}{UTF8}{gbsn}`阿卡"\end{CJK}. MM-GAN outperforms all baselines, information in additional attribute tags such as \begin{CJK}{UTF8}{gbsn}``复古$|$ (retro)", ``Artka"\end{CJK}), and other information from the product main image are together considered by the model and help the model select core words and filter out irrelevant words in generated product short titles. It shows that MM-GAN using different types of inputs can help generate better product short titles. To leave a deeper impression on the performance of our proposed model, we also put more online samples generated by the MM-GAN in a module of the TaoBao App, as shown in Fig. \ref{fig:example-demo}. From all generated samples we also find few bad cases which are not shown online (e.g., repetitive words in the generated short titles, wrong generated words which are not related to the product at all), leaving a great space for further improvement.

%% file: conclusion.tex
\section{Conclusion}

In this paper, we propose a multi-modal generative adversarial network for short product title generation in E-commerce. Different from conventional methods which only consider textual information from long product titles, we design a multi-modal generative model to incorporate additional information from product image and attribute tags.
% Besides, we focus on overall quality of the generated short titles and evaluate them in a human-like view.
% Experimental results verify the effectiveness of the proposed method on a large E-commerce dataset and the online deployment in a real environment of an E-commerce app. 
% Experimental results verify the effectiveness of the proposed method on a large E-commerce dataset
% , and the online deployment in a real environment of an E-commerce app shows that our method improves the click through rate by 1.66\% and click conversion rate by 1.87\%, respectively.
Extensive experiments on a large real-world E-commerce dataset verify the effectiveness of our proposed model when comparing with several state-of-the-art baselines. 
Moreover, the online deployment in a real environment of an E-commerce app shows that our method can improve the click through rate and click conversion rate.

\section{Acknowledgement}
This work is supported in part by NSF through grants IIS-1526499, IIS-1763325, and CNS-1626432, and NSFC 61672313. We thank the anonymous reviewers for their valuable comments.

%% file: main.bbl
\begin{thebibliography}{33}
\expandafter\ifx\csname natexlab\endcsname\relax\def\natexlab#1{#1}\fi

\bibitem[{Antol et~al.(2015)Antol, Agrawal, Lu, Mitchell, Batra,
  Lawrence~Zitnick, and Parikh}]{antol2015vqa}
Stanislaw Antol, Aishwarya Agrawal, Jiasen Lu, Margaret Mitchell, Dhruv Batra,
  C~Lawrence~Zitnick, and Devi Parikh. 2015.
\newblock Vqa: Visual question answering.
\newblock In \emph{ICCV}.

\bibitem[{Bachman and Precup(2015)}]{bachman2015data}
Philip Bachman and Doina Precup. 2015.
\newblock Data generation as sequential decision making.
\newblock In \emph{NIPS}.

\bibitem[{Bahdanau et~al.(2015)Bahdanau, Cho, and Bengio}]{bahdanau2014neural}
Dzmitry Bahdanau, Kyunghyun Cho, and Yoshua Bengio. 2015.
\newblock Neural machine translation by jointly learning to align and
  translate.
\newblock In \emph{ICLR}.

\bibitem[{Cao et~al.(2016)Cao, Li, Li, Wei, and Li}]{cao2016attsum}
Ziqiang Cao, Wenjie Li, Sujian Li, Furu Wei, and Yanran Li. 2016.
\newblock Attsum: Joint learning of focusing and summarization with neural
  attention.
\newblock In \emph{COLING}.

\bibitem[{Chen et~al.(2016)Chen, Zhu, Ling, Wei, and
  Jiang}]{chen2016distraction}
Qian Chen, Xiaodan Zhu, Zhenhua Ling, Si~Wei, and Hui Jiang. 2016.
\newblock Distraction-based neural networks for document summarization.
\newblock In \emph{IJCAI}.

\bibitem[{Chopra et~al.(2016)Chopra, Auli, and Rush}]{chopra2016abstractive}
Sumit Chopra, Michael Auli, and Alexander~M Rush. 2016.
\newblock Abstractive sentence summarization with attentive recurrent neural
  networks.
\newblock In \emph{NAACL-HLT}.

\bibitem[{Clarke and Lapata(2008)}]{clarke2008global}
James Clarke and Mirella Lapata. 2008.
\newblock Global inference for sentence compression: An integer linear
  programming approach.
\newblock \emph{Journal of Artificial Intelligence Research}.

\bibitem[{Dai et~al.(2017)Dai, Fidler, Urtasun, and Lin}]{dai2017towards}
Bo~Dai, Sanja Fidler, Raquel Urtasun, and Dahua Lin. 2017.
\newblock Towards diverse and natural image descriptions via a conditional gan.
\newblock In \emph{ICCV}.

\bibitem[{Fedus et~al.(2018)Fedus, Goodfellow, and Dai}]{fedus2018maskgan}
William Fedus, Ian Goodfellow, and Andrew~M Dai. 2018.
\newblock Maskgan: Better text generation via filling in the \_.
\newblock In \emph{ICLR}.

\bibitem[{Gong et~al.(2018)Gong, Luo, Zhu, Liu, and Ou}]{gong2018automatic}
Yu~Gong, Xusheng Luo, Kenny~Q Zhu, Shichen Liu, and Wenwu Ou. 2018.
\newblock Automatic generation of chinese short product titles for mobile
  display.
\newblock In \emph{IAAI}.

\bibitem[{Goodfellow et~al.(2014)Goodfellow, Pouget-Abadie, Mirza, Xu,
  Warde-Farley, Ozair, Courville, and Bengio}]{goodfellow2014generative}
Ian Goodfellow, Jean Pouget-Abadie, Mehdi Mirza, Bing Xu, David Warde-Farley,
  Sherjil Ozair, Aaron Courville, and Yoshua Bengio. 2014.
\newblock Generative adversarial nets.
\newblock In \emph{NIPS}.

\bibitem[{Guo et~al.(2017)Guo, Lu, Cai, Zhang, Yu, and Wang}]{guo2017long}
Jiaxian Guo, Sidi Lu, Han Cai, Weinan Zhang, Yong Yu, and Jun Wang. 2017.
\newblock Long text generation via adversarial training with leaked
  information.
\newblock In \emph{AAAI}.

\bibitem[{Hochreiter and Schmidhuber(1997)}]{hochreiter1997long}
Sepp Hochreiter and J{\"u}rgen Schmidhuber. 1997.
\newblock Long short-term memory.
\newblock \emph{Neural computation}.

\bibitem[{Isola et~al.(2017)Isola, Zhu, Zhou, and Efros}]{isola2017image}
Phillip Isola, Jun-Yan Zhu, Tinghui Zhou, and Alexei~A Efros. 2017.
\newblock Image-to-image translation with conditional adversarial networks.
\newblock In \emph{ICCV}.

\bibitem[{Kingma and Ba(2015)}]{kingma2014adam}
Diederik~P Kingma and Jimmy Ba. 2015.
\newblock Adam: A method for stochastic optimization.
\newblock In \emph{ICLR}.

\bibitem[{Lamb et~al.(2016)Lamb, GOYAL, Zhang, Zhang, Courville, and
  Bengio}]{lamb2016professor}
Alex~M Lamb, Anirudh Goyal ALIAS~PARTH GOYAL, Ying Zhang, Saizheng Zhang,
  Aaron~C Courville, and Yoshua Bengio. 2016.
\newblock Professor forcing: A new algorithm for training recurrent networks.
\newblock In \emph{NIPS}.

\bibitem[{Ledig et~al.(2017)Ledig, Theis, Husz{\'a}r, Caballero, Cunningham,
  Acosta, Aitken, Tejani, Totz, Wang et~al.}]{ledig2017photo}
Christian Ledig, Lucas Theis, Ferenc Husz{\'a}r, Jose Caballero, Andrew
  Cunningham, Alejandro Acosta, Andrew~P Aitken, Alykhan Tejani, Johannes Totz,
  Zehan Wang, et~al. 2017.
\newblock Photo-realistic single image super-resolution using a generative
  adversarial network.
\newblock In \emph{CVPR}.

\bibitem[{Li et~al.(2017)Li, Monroe, Shi, Jean, Ritter, and
  Jurafsky}]{li2017adversarial}
Jiwei Li, Will Monroe, Tianlin Shi, S{\'e}bastien Jean, Alan Ritter, and Dan
  Jurafsky. 2017.
\newblock Adversarial learning for neural dialogue generation.
\newblock In \emph{EMNLP}.

\bibitem[{Lin(2004)}]{lin2004rouge}
Chin-Yew Lin. 2004.
\newblock Rouge: A package for automatic evaluation of summaries.
\newblock \emph{Text Summarization Branches Out}.

\bibitem[{Lin et~al.(2017)Lin, Li, He, Zhang, and Sun}]{lin2017adversarial}
Kevin Lin, Dianqi Li, Xiaodong He, Zhengyou Zhang, and Ming-Ting Sun. 2017.
\newblock Adversarial ranking for language generation.
\newblock In \emph{NIPS}.

\bibitem[{Miao and Blunsom(2016)}]{miao2016language}
Yishu Miao and Phil Blunsom. 2016.
\newblock Language as a latent variable: Discrete generative models for
  sentence compression.
\newblock In \emph{EMNLP}.

\bibitem[{Mihalcea(2005)}]{mihalcea2005language}
Rada Mihalcea. 2005.
\newblock Language independent extractive summarization.
\newblock In \emph{ACL}.

\bibitem[{Nallapati et~al.(2017)Nallapati, Zhai, and
  Zhou}]{nallapati2017summarunner}
Ramesh Nallapati, Feifei Zhai, and Bowen Zhou. 2017.
\newblock Summarunner: A recurrent neural network based sequence model for
  extractive summarization of documents.
\newblock In \emph{AAAI}.

\bibitem[{Ranzato et~al.(2016)Ranzato, Chopra, Auli, and
  Zaremba}]{ranzato2015sequence}
Marc'Aurelio Ranzato, Sumit Chopra, Michael Auli, and Wojciech Zaremba. 2016.
\newblock Sequence level training with recurrent neural networks.
\newblock In \emph{ICLR}.

\bibitem[{See et~al.(2017)See, Liu, and Manning}]{see2017get}
Abigail See, Peter~J Liu, and Christopher~D Manning. 2017.
\newblock Get to the point: Summarization with pointer-generator networks.
\newblock In \emph{ACL}.

\bibitem[{Simonyan and Zisserman(2015)}]{simonyan2014very}
Karen Simonyan and Andrew Zisserman. 2015.
\newblock Very deep convolutional networks for large-scale image recognition.
\newblock In \emph{ICLR}.

\bibitem[{Sutskever et~al.(2014)Sutskever, Vinyals, and
  Le}]{sutskever2014sequence}
Ilya Sutskever, Oriol Vinyals, and Quoc~V Le. 2014.
\newblock Sequence to sequence learning with neural networks.
\newblock In \emph{NIPS}.

\bibitem[{Wan et~al.(2018)Wan, Zhao, Yang, Xu, Ying, Wu, and
  Yu}]{wan2018improving}
Yao Wan, Zhou Zhao, Min Yang, Guandong Xu, Haochao Ying, Jian Wu, and Philip~S
  Yu. 2018.
\newblock Improving automatic source code summarization via deep reinforcement
  learning.
\newblock In \emph{Proceedings of the 33rd ACM/IEEE International Conference on
  Automated Software Engineering}, pages 397--407. ACM.

\bibitem[{Wang et~al.(2018)Wang, Tian, Qiu, Li, Lang, Si, and
  Lan}]{wang2018multi}
Jingang Wang, Junfeng Tian, Long Qiu, Sheng Li, Jun Lang, Luo Si, and Man Lan.
  2018.
\newblock A multi-task learning approach for improving product title
  compression with user search log data.
\newblock In \emph{AAAI}.

\bibitem[{Wang and Wan(2018)}]{wang2018sentigan}
Ke~Wang and Xiaojun Wan. 2018.
\newblock Sentigan: Generating sentimental texts via mixture adversarial
  networks.
\newblock In \emph{IJCAI}.

\bibitem[{Williams(1992)}]{williams1992simple}
Ronald~J Williams. 1992.
\newblock Simple statistical gradient-following algorithms for connectionist
  reinforcement learning.
\newblock \emph{Machine learning}.

\bibitem[{Woodsend and Lapata(2010)}]{woodsend2010automatic}
Kristian Woodsend and Mirella Lapata. 2010.
\newblock Automatic generation of story highlights.
\newblock In \emph{ACL}.

\bibitem[{Yu et~al.(2017)Yu, Zhang, Wang, and Yu}]{yu2017seqgan}
Lantao Yu, Weinan Zhang, Jun Wang, and Yong Yu. 2017.
\newblock Seqgan: Sequence generative adversarial nets with policy gradient.
\newblock In \emph{AAAI}.

\end{thebibliography}
